%% file: Main.tex
\documentclass[conference]{IEEEtran}
\IEEEoverridecommandlockouts

\ifCLASSINFOpdf
   \usepackage[pdftex]{graphicx}
 
\else
 
\fi

\usepackage{microtype}

\usepackage{amsmath}
\usepackage{tikz}
\usepackage{mathdots}
\usepackage{yhmath}
\usepackage{cancel}
\usepackage{color}
\usepackage{siunitx}
\usepackage{array}
\usepackage{multirow}
\usepackage{amssymb}
\usepackage{gensymb}
\usepackage{tabularx}
\usepackage{extarrows}
\usepackage{booktabs}
\usetikzlibrary{fadings}
\usetikzlibrary{patterns}
\usetikzlibrary{shadows.blur}
\usetikzlibrary{shapes}
\usepackage{lipsum,adjustbox}
\usepackage{amsfonts}
\usepackage{algorithm}
\usepackage{algpseudocode}
\newtheorem{lemma}{Lemma}

\newtheorem{remark}{Remark}

\usepackage[utf8]{inputenc} % 'cp1252'-Western, 'cp1251'-Cyrillic, etc.
\usepackage[english]{babel} % 'french', 'german', 'spanish', 'danish', etc.

\usepackage[classicReIm]{kpfonts}
\usepackage{graphicx}

\begin{document}
%

% \title{Nussbaum Gain Adaptive Neural Control for Optimized Aircraft spin Recovery Maneuvers}

\title{Nussbaum Function Based Approach for Tracking Control of Robot Manipulators }

\author{\IEEEauthorblockN{Hamed Rahimi Nohooji}
\IEEEauthorblockA{Interdisciplinary Centre for Security, Reliability\\ and Trust (SnT),
Automation Robotics Research Group\\
University of Luxembourg\\
Email: hamed.rahimi@uni.lu}
\and
\IEEEauthorblockN{Holger Voos}
\IEEEauthorblockA{Interdisciplinary Centre for Security, Reliability\\ and Trust (SnT),
Automation Robotics Research Group\\
\IEEEauthorblockA{Faculty of Science, Technology and Medicine (FSTM)\\
Department of Engineering\\
University of Luxembourg}
Email: holger.voos@uni.lu}
}

\maketitle
\IEEEpeerreviewmaketitle

\input{Robot/00-Abstract}

% \begin{IEEEkeywords}
% Nussbaum Gain, Aircraft spin Recovery, PID control, Neural network, Adaptive Control
% \end{IEEEkeywords}


% \input{Aircraft/01-Intro}
% % \input{Aircraft/02-Related}
% \input{Aircraft/02-Dynamic}
% \input{Aircraft/03-Control}
%  \input{Aircraft/04-Simulation}
%  \input{Aircraft/05-Experiment}
%  \input{Aircraft/06-Disussion}
%  \input{Aircraft/07-Conclusion}

\input{Robot/01-Intro}

\input{Robot/03-Dyn.tex}

\input{Robot/04-Control}

 \input{Robot/05-Simulation}

 % % \input{Robot/05-Experiment}
 % \input{Robot/06-Discuss}
 \input{Robot/07-Conclusion}

\section*{Acknowledgment}
The authors would like to acknowledge the support from the Luxembourg National Research Fund (FNR) under the project IC22/IS/17432865/COSAMOS.

% This work is supported by FNR "Fonds national de
% la Recherche" (Luxembourg) through XXX (ref. XXX).

 \bibliographystyle{IEEEtran}
\bibliography{refs.bib}

\end{document}

%% file: Robot/00-Abstract.tex
\begin{abstract}

This paper introduces a novel Nussbaum function-based PID control for robotic manipulators. The integration of the Nussbaum function into the PID framework provides a solution with a simple structure that effectively tackles the challenge of unknown control directions.
 Stability is achieved through a combination of neural network-based estimation and Lyapunov analysis, facilitating automatic gain adjustment without the need for system dynamics. Our approach offers a gain determination with minimum parameter requirements, significantly reducing the complexity and enhancing the efficiency of robotic manipulator control. 
The paper guarantees that all signals within the closed-loop system remain bounded. Lastly, numerical simulations validate the theoretical framework, confirming the effectiveness of the proposed control strategy in enhancing robotic manipulator control.

% This paper introduces a novel low-complexity Nussbaum Function-Based PID control for robot manipulators, addressing the challenge of unknown dynamics and control directions. By integrating the adaptability of the Nussbaum function within a PID framework, we propose a solution that simplifies the control design process while ensuring system stability and performance. Our approach significantly reduces the need for parameter tuning by utilizing linking parameters and guarantees system stability through direct Lyapunov analysis. The efficacy of the proposed control strategy is validated through analytical analysis, and numerical simulations, offering a cost-effective and straightforward solution for enhancing robotic manipulator control.

\end{abstract}

%% file: Robot/01-Intro.tex
\section{Introduction}
\label{sec_intro}

The rapid expansion of robotic systems across various industries has driven the development of advanced control mechanisms to enhance their functionality and adaptability.
Despite the progress facilitated by rigorous mathematical frameworks, the nonlinearities and uncertainties intrinsic to robotic operations continue to pose significant challenges. This reality highlights the need for advanced control strategies like fuzzy logic \cite{yilmaz2021self}, Kalman filter \cite{lightcap2009extended},  iterative learning control \cite{zhu2020estimation}, and actor–critic learning \cite{nohooji2024actor} to effectively counteract these complexities. However, the inherent complexity of these methods often complicates their application in real-world scenarios, underscoring the ongoing research imperative to refine and simplify control solutions for robotic manipulators.

In the field of control for robotic manipulators, PID control is recognized for its intuitive design and simplicity, positioning it as a pillar for both theoretical exploration and practical application across numerous real-world systems \cite{cervantes2001pid, borase2021review}. It addresses the widespread issue of complexity in robotic control strategies. However, traditional PID controls face limitations, particularly in weight updating, stability assurance, and the need for extensive parameter tuning in dynamic environments \cite{ajwad2015systematic}. Attempts to refine PID control, such as employing optimization techniques for gain tuning, aim to enhance its adaptability \cite{armendariz2014neuro, belkadi2017robust, azeez2023optimization}. However, these enhancements can introduce new complexities, somewhat undermining the original appeal of PID control's simplicity.

Understanding the direction of control gain is crucial in robotic system control, as incorrect application can destabilize the system instead of guiding it to the desired state. While various methods exist to address the unknown control direction, such as logic-based switching \cite{huang2018tuning}, extreme seeking \cite{scheinker2012minimum}, and nonlinear PI control \cite{psillakis2016consensus}, the Nussbaum function approach \cite{nussbaum1983some} is the most studied method \cite{zhao2022adaptive, song2017robust, habibi2018adaptive, habibi2019backstepping}. Implementing Nussbaum-based strategies in robotic control is effective; however, their integration into the comprehensive control framework can complicate the overall system. This scenario highlights the importance of developing more accessible strategies that utilize the Nussbaum function's advantages while avoiding excessive complexity in the control framework.

Given the complexities and challenges previously discussed, our research is driven by the need to develop a control strategy that is simple and low in complexity, effectively managing the critical aspect of unknown control gain direction in robotic manipulators. Acknowledging the limitations of traditional PID controls and the complexities of Nussbaum-based methods, we introduce a novel approach that combines the simple structure of PID control with the capability of the Nussbaum function. This method ensures stability and enhances performance, offering a control mechanism that is inexpensive in online computational demands, and effectively bridging the mentioned gap in robotic control strategies.

In this work, we present a control strategy for robotic manipulators, tackling unknown dynamics and control directions through an adaptive Nussbaum-based control. Our approach, rooted in the PID control structure, simplifies yet enhances the control scheme's effectiveness. Different from conventional PID methods, we ensure the closed-loop system's stability through direct Lyapunov analysis. Additionally, our strategy features automatic gain adjustment and utilizes linking parameters, significantly reducing the number of tuning requirements, and thereby enhancing the system's efficiency and response without complicating the control framework.
The main contribution can be summarized as follows:

\begin{itemize}

    \item We integrate the Nussbaum function into the PID structure for the control of robotic manipulators. The controller is simple, yet it guarantees stability without requiring knowledge of robot dynamics or control direction.
   
    \item By employing linked PID terms and establishing adaptive laws, the controller automates gain determination with minimal parameter requirements, thereby facilitating its application in real-time scenarios.

    % \item The developed control scheme was analytically analyzed, and its effectiveness was verified through numerical simulations.
\end{itemize}

The rest of the paper is structured as follows: Section II delves into problem formulation and preliminaries, setting the theoretical groundwork. Section III introduces the Nussbaum Function-based control design and its stability analysis. Numerical simulations demonstrating the control strategy's effectiveness are presented in Section IV. The paper concludes with Section V, summarizing the key findings and implications of this research.

%% file: Robot/03-Dyn.tex
\section{Problem Formulation And Priliminaries}
\label{sec_Dyn}

\subsection{Problem Formulation}
Consider the robotic system described in the joint space \( q(t) \in \mathbb{R}^{n \times 1} \) as \cite{nohooji2018neural, chen2016adaptive},
\begin{equation} \label{eq_dyn}
M(q(t))\ddot{q}(t) + C(q(t),\dot{q}(t))\dot{q}(t) + G(t) = \tau(t) \quad 
\end{equation}
where \( M(q(t)) \in \mathbb{R}^{n \times n} \) denotes a positive definite inertial matrix; \( C(q(t),\dot{q}(t)) \in \mathbb{R}^{n \times n} \) represents the Coriolis and centrifugal matrix; \( G(t) \in \mathbb{R}^{n \times 1} \) denotes the gravitational force vector. 
 \( \tau(t) \in \mathbb{R}^{n \times 1} \) represents the actual torque applied to the robotic system. 
Furthermore, in practical systems, the actual torque is affected by unknown actuator dynamics and can be expressed as \cite{chen2015saturated},
\begin{equation}
\tau(t) = \kappa u(t), \quad 
\end{equation}
where the nonlinear matrix \( \kappa \in \mathbb{R}^{n \times n} \) is an unknown control direction, and \( u(t) \) represents the control input.\\
The following properties of robot manipulators dynamics (\ref{eq_dyn}) are required for control analysis \cite{lewis2003robot},
\begin{itemize}
\item \textbf{P 1:} \( M(q(t)) \) is symmetric positive definite.
\item \textbf{P 2:} \( \frac{1}{2}(C(q(t),\dot{q}(t))) - \dot{M}(q(t))) \) is skew-symmetric.
\item \textbf{P 3:} There exist unknown positive constants $m_-$, $m^+$, $c$ and $g$ such that $m_- \leq  M(q(t)) \leq m^+$,  $ \|C(q(t),\dot{q}(t))\| \leq c \|\dot{q}(t)\|,$ and $\|G(q(t))\| \leq g < \infty.
$
\end{itemize}

The goal of this paper is to design a control law for a robot dynamics (1) such that 
\begin{itemize}
    \item  \textbf{G 1:} the closed-loop system is stable and all the signals remain uniformly bounded.
    
    \item \textbf{G 2:} the joint position signal \( q(t) \) closely tracks a specified desired trajectory \( q_d(t) \), so that the limit \(\lim_{t \to \infty} |q_i(t) - q_{di}(t)| = \eta_i\), for $i=1,2,...n$, and \(\eta_i\) being a small positive constant.

\end{itemize}

To this end, the following assumptions are needed.

% Assumptions
% Assumption 1
\newtheorem{assumption}{Assumption}
\begin{assumption}
\label{assumption:bounds}
We consider each link of the manipulator as a slender rod with uniform mass distribution, ignoring additional components' mass and inertia.
\end{assumption}

\begin{assumption}
The desired trajectory \( q_d(t) \), along with its first \( \dot{q}_d(t) \) and second derivatives \( \ddot{q}_d(t) \), are assumed to be smooth, known functions of time and are bounded. Additionally, the robot position vector \( q \) is assumed to be accessible for the purposes of control design.
\end{assumption}

\begin{remark}
The dynamics of the system, as described by equation (1), are fully unknown, which presents a significant challenge in the control design. This paper addresses the inherent difficulty of unknown control direction by employing a Nussbaum-type function. The Nussbaum approach is particularly adept at handling systems with unknown control coefficients, adapting to the control direction without prior knowledge of the system's dynamics. 
\end{remark}

\subsection{Priliminaries}

To approximate the system uncertainties including the manipulator's unknown dynamics, in its continuous 
movement, we utilize linear-in-parameter approximators with an a priori-defined basis function vector and a yet-to-be-learned unknown parameter vector \cite{pane2019reinforcement}. To this purpose, a generic function approximator $F(x)$ is denoted by 
$
F(x, \psi) = \psi^\top \phi(x),
$
where $\psi \in \mathbb{R}^{n_p}$ is the unknown parameter vector of dimension $n_p$ and $\phi(x) \in \mathbb{R}^{n_p}$ is the user-defined known basis function vector with ${n_p}$ being the NN input dimension, and 
 $ x \in \mathbb{R}^{m}$ is the neural network input vector with $m$ being the neural network input dimension.
In this work, we have used the radial basis function (RBF) given by $
\tilde{\phi}(x) = e^{-0.5(x-c)^\top B^{-1}(x-c)},$
where $c \in \mathbb{R}^n$ is the center and $B \in \mathbb{R}^{n \times n}$ the covariance matrix of the RBF. 

Define the position error, \( e(t) \in \mathbb{R}^n \), as \( e(t) = q_d(t) -q(t) \) and the velocity error, \( \dot{e}(t) \in \mathbb{R}^n \), as \( \dot{e}(t) =  \dot{q}_d(t) - \dot{q}(t)\). To further our analysis, we define the generalized intermediate variable \( \Psi(t) \) as follows:
\begin{equation} \label{eq:E}
\Psi(t) = 2\gamma e(t) + \gamma^2 \int_0^t e(\rho) \, d\rho + \frac{d}{dt}e(t),
\end{equation}
where \( \gamma > 0 \). This formulation allows us to address the tracking error by stabilizing \( \Psi(t) \) using the following lemma.

\begin{lemma} \cite{chen2021tracking}
Given the intermediate variable \( \Psi(t) \) as defined in \eqref{eq:E}, if \( \Psi(t) \to 0 \) as \( t \to \infty \), then the tracking errors \( e(t) \) and \( \dot{e}(t) \), and their integrals are bounded and converge to zero over time.
\end{lemma}

\newtheorem{definition}{Definition}
\begin{definition}[Nussbaum Function]
A Nussbaum function \(N(\zeta)\) is characterized by its capability to handle unknown control directions within a control system. For a continuously differentiable function \(N(\zeta) : [0, \infty) \to (-\infty, \infty)\), it is defined via its positive and negative truncated forms, \(N^+(\zeta) = \max\{0, N(\zeta)\}\) and \(N^-(\zeta) = \max\{0, -N(\zeta)\}\), respectively. The function satisfies the condition that
\begin{equation}
N(\zeta) = N^+(\zeta) - N^-(\zeta),
\end{equation}
with the properties that for any \(\zeta\),
\begin{equation}
\lim_{v \to \infty} \sup \frac{1}{v} \left[ \int_0^v N^+(\zeta) d\zeta - \int_0^v N^-(\zeta) d\zeta \right] = \infty,
\end{equation}
\begin{equation}
\lim_{v \to \infty} \sup \frac{1}{v} \left[ \int_0^v N^-(\zeta) d\zeta - \int_0^v N^+(\zeta) d\zeta \right] = \infty.
\end{equation}
A function that satisfies these conditions is utilized for managing the uncertainty in control direction, offering a robust approach to control system design.
\end{definition}

\begin{lemma} \label{Lemma2} \cite{song2017robust, habibi2018adaptive} 
Let \(V(t)\) and \(\zeta(t)\) be smooth functions defined over the interval \([0, t_f)\) with \(V(t) > 0\) for all \(t \in [0, t_f)\). Given \(N(\zeta(t))\) as a Nussbaum-type function, if for any \(t \in [0, t_f)\), the following condition is met:
\begin{equation} \label{eq:boundedness_nussbaum}
V(t) < c_0 + e^{-c_1 t} \int_0^t \left(gN(\zeta(\tau)) + 1\right) \dot{\zeta}(\tau) e^{c_1 \tau} d\tau,
\end{equation}
where \(c_0 > 0\) and \(c_1 > 0\) are positive constants, and \(g\) is a control parameter within closed intervals \(L = [l_-, l_+]\) excluding zero (\(0 \notin L\)), then \(V(t)\), \(\zeta(t)\), and the integral \(\int_0^t g N(\zeta(\tau))\dot{\zeta}(\tau)e^{c_1 \tau} d\tau\) are guaranteed to be bounded on \([0, t_f)\).
\end{lemma}

%% file: Robot/04-Control.tex
\section{Nussbaum Function-based Control Design and Stability Analysis}
\label{sec_control}
\noindent 

We propose a Nussbaum function based PID-like control input for robot manipulator control, delineated as:
\begin{equation} \label{eq:control_law}
\begin{aligned}
u(t) = &\left(k_{\pi} + \kappa_{\pi}(t)\right) K_N(\zeta)e(t) + \left(k_{\iota} + \kappa_{\iota}(t)\right)K_N(\zeta)\int_0^t e(\rho) \, d\rho \\
&+ \left(k_{\Delta} + \kappa_{\Delta}(t)\right)K_N(\zeta)\frac{d}{dt}e(t),
\end{aligned}
\end{equation}

\noindent
with $K_N(\zeta)=-N(\zeta)$ is the related Nussbaum function  gain.
In \eqref{eq:control_law} we employed constant gains \(k_{\pi}\), \(k_{\iota}\), and \(k_{\Delta}\) alongside their time-varying analogs \(\kappa_{\pi}(t)\), \(\kappa_{\iota}(t)\), and \(\kappa_{\Delta}(t)\).  This strategy enhances traditional PID controls by integrating adaptive gains that adjust in real-time to the system's state. To further simplify parameter tuning, we establish the relationships \(k_{\Delta} = \frac{k_{\pi}}{2\gamma} = \frac{k_{\iota}}{\gamma^2}\) and \(\kappa_{\Delta}(t) = \frac{\kappa_{\pi}(t)}{2\gamma} = \frac{\kappa_{\iota}(t)}{\gamma^2}\), ensuring the quadratic expression \(S^2 + 2\gamma S + \gamma^2\) is Hurwitz polynomial, where $S$ is the Laplace operator. Thus, the control design primarily depends on selecting \(k_{\Delta}\) and \(\kappa_{\Delta}(t)\). The PID-like control input is then refined to:
\begin{equation} \label{eq:simplified_control}
\begin{aligned}
u(t) = &\left(k_{\Delta} + \kappa_{\Delta}(t)\right)K_N(\zeta)\left(2\gamma e(t) + \gamma^2 \int_0^t e(\rho) \, d\rho + \frac{d}{dt}e(t)\right)\\
= &-\left(k_{\Delta} + \kappa_{\Delta}(t)\right) N(\zeta)\Psi.
\end{aligned}
\end{equation}
This approach significantly reduces the complexity of the gain selection process, focusing on only two key parameters. 
The time-varying gain \(\kappa_{\Delta}(t)\) is adaptively updated by
\begin{equation} \label{kappa}
\kappa_{\Delta}(t) = -\alpha \hat{\psi}(t)^T \phi(x),
\end{equation}
and the adaptive law for \(\hat{\psi}(t)\) is given by
\begin{equation} \label{eq:hat{psi}}
\dot{\hat{\psi}}(t) = -\Gamma(\alpha\|\Psi(t)\|^2 \phi(x) + \sigma\hat{\psi}(t)).
\end{equation}
with $\alpha$ and $\phi$ are positive control constants, and $\Gamma=\Gamma^T>0$.

In this work, we choose $ N(\zeta) = \zeta^2 \cos(\zeta)$ with the property of $N(0)=0$
as our Nussbaum function. The updating law for $\zeta$ is given by
\begin{equation} \label{eq:dot{zeta}}
\dot{\zeta}(t) = \Psi(t)^T(k_\Delta + \kappa_\Delta(t))\Psi(t).
\end{equation}

The above control framework, enhanced by adaptive laws based on the low complexity Nussbaum function PID control, provides us with an effective approach to robotic manipulator control, as formalized in the forthcoming theorem.

 \newtheorem{theorem}{Theorem}
\begin{theorem}
Consider the robot manipulator system described by (1) with the Properties 1-3, satisfying to Assumptions 1 and 2. If the Nussbaum function-based PID-like control law \eqref{eq:simplified_control}, with the adaptive updating laws \eqref{eq:hat{psi}}, and \eqref{eq:dot{zeta}}, and and Lemmas 1, and 2 is implemented, then, with the design parameters are properly chosen, the closed-loop system remains stable, and all signals within the system are uniformly bounded. Furthermore, the joint position signal \(q(t)\) closely tracks the desired trajectory \(q_d(t)\), with the tracking error converging to a small neighborhood around zero. 
\end{theorem}

\newenvironment{proof}{{\bfseries Proof:}}{\hfill$\square$}
\begin{proof}
Consider the Lyapunov candidate as
\begin{equation} \label{Lyap}
V(t) = \frac{1}{2} \Psi(t)^T M(q(t)) \Psi(t) + \frac{1}{2} \Tilde{\psi}(t)^T \Gamma^{-1} \Tilde{\psi}(t),
\end{equation}
where \(\Tilde{\psi}(t) = \hat{\psi}(t) - \psi^*\) with $\psi^*(t)$ is the ideal constant weight vector of neural network approximation.

\noindent
The time derivative of the Lyapunov function $V(t)$ can be bounded as
\begin{equation} \label{eq_vd}
{\dot V(t)} = {\Psi(t) ^T}M(t)\dot \Psi(t)  + \frac{1}{2}{\Psi(t) ^T}\dot M(t)\Psi(t)+   \Tilde{\psi}(t)^T \Gamma^{-1} \dot{\hat{\psi}}(t).
\end{equation}

\noindent
Considering dynamics (1) and the definition of the generalized error \(\Psi(t)\), one can obtain \(M(t)\dot{\Psi}(t) = C(t)\dot{q}(t) + G(t) - \tau(t) + M(t)\Bar{e}(\cdot)\), where \(\Bar{e}(\cdot) = \ddot{q}_d(t) + 2\gamma \dot{e}(t) + \gamma^2 e(t)\). Then, utilizing the application of Young’s inequality, and considering Property 2, we obtain \(\Psi(t)^T C(t) \dot{q}(t) \leq \alpha \|\Psi(t)\|^2 c^2 \|\dot{q}(t)\|^4 + \frac{1}{4\alpha}\), \(\Psi(t)^T G(t) \leq \alpha \|\Psi(t)\|^2 g^2 + \frac{1}{4\alpha}\), \(\Psi(t)^T M(t) \Bar{e}(\cdot) \leq \alpha \|\Psi(t)\|^2 m^{+^2} \|\Bar{e}(\cdot)\|^2 + \frac{1}{4\alpha}\), and \(\frac{1}{2} \Psi(t)^T \dot{M}(t) \Psi(t) = \Psi(t)^T C(t) \Psi(t) \leq \alpha \|\Psi(t)\|^2 c^2 \|\dot{q}(t)\|^2 \|\Psi(t)\|^2 + \frac{1}{4\alpha}\), where \(\alpha > 0\) is a design parameter. Accordingly, taking into account equation \(\eqref{eq_vd}\) and applying the aforementioned inequalities, one has,
\begin{equation} \label{eq:vdt}
\dot{V}(t) \leq \alpha \|\Psi(t)\|^2 \Lambda(t) - \Psi(t)^T \kappa u(t) + \frac{1}{\alpha} + \Gamma^{-1} \Tilde{\psi}(t)^T \dot{\hat{\psi}}(t),
\end{equation}
where 

$\Lambda(t) = -\left(c^2\|\dot{q}(t)\|^4 + g^2+ m^{+2}\|\Bar{e}(t)\|^2 + c^2 \|\dot{q}(t)\|^2 \|\Psi(t)\|^2\right)$.

\noindent
Subsequently, considering  neural networks approximation $
\Lambda(t) = \psi^* \phi(x)+\varepsilon(x)
$, with $\varepsilon \left(x\right)$ being the unknown approximation error which is upper bounded in the sense that $\left\| \varepsilon \left(x\right)\right\| \le \varepsilon _{m}$, and
utilizing \eqref{eq:hat{psi}}, \eqref{eq:dot{zeta}}, and \eqref{eq:simplified_control}  into (\ref{eq:vdt}), and considering $-\Tilde{\psi}(t)^T {\hat{\psi}}(t) \le - \Tilde{\psi}(t)^T\Tilde{\psi}(t) + |\psi^*\|^2$ \cite{nohooji2024actor},  yields
\begin{align*}
\dot{V}(t) &\leq \dot{\zeta}(t) - \|\Psi(t)\|^2(k_\Delta + \kappa_\Delta(t)) - \alpha \|\Psi(t)\|^2(\psi^{*^\top} \phi(x) + \varepsilon) \\
&+ \Psi(t)^T \kappa N(\zeta(t))(k_\Delta + \kappa_\Delta(t)) \Psi(t) + \frac{1}{\alpha}  \\
&- \alpha \|\Psi(t)\|^2 \Tilde{\psi}(t)^T \phi(x) - \frac{1}{2} \sigma\Tilde{\psi}(t)^T\Tilde{\psi}(t) + \frac{1}{2} \sigma \|\psi^{*}\|^2.
\end{align*}
Considering \(\|\Psi(t)\|^2 \kappa_{\Delta}(t) = -\|\Psi(t)\|^2 \alpha \hat{\psi}(t)^T \phi(x)\), the above equation can be rewritten as
\begin{align*}
\dot{V}(t) &\leq \dot{\zeta}(t) + \kappa N(\zeta(t)) \dot{\zeta}(t) - \|\Psi(t)\|^2 (k_{\Delta} + \alpha \varepsilon)\\
&- \frac{1}{2} \sigma\ \|\Tilde{\psi}(t)\|^2 + \frac{1}{\alpha} + \frac{1}{2} \sigma \Bar{\psi}^2,
\end{align*}
with \(\Bar{\psi}\) being the upper bound of the optimal weight \(\|\psi^{*}\|\).

\noindent
Finally, the above equation can be formed as 
\begin{equation} \label{eq22}
\dot{V}(t) < \dot{\zeta}(t) + \kappa N(\zeta(t)) \dot{\zeta}(t) - c_1 V(t) + c_2, 
\end{equation}
where, 
$c_1 = \min\left(\frac{2(\alpha\varepsilon_m + k_{\Delta})}{m^+}, \frac{\sigma}{\lambda_{\max}(\Gamma^{-1})}\right)
$
and $c_2 = \frac{1}{\alpha}+\frac{1}{2} \sigma \Bar{\psi}^2$ and both are positive. Multiplying  \eqref{eq22} by \(e^{c_1 t} > 0\), yields
\begin{equation} \label{eq:24}
\frac{d}{dt}\left( V(t) e^{c_1 t} \right) \leq \dot{\zeta}(t)e^{c_1 t} + \kappa N(\zeta (t))\dot{\zeta}(t)e^{c_1 t} + c_2 e^{c_1 t},
\end{equation}
and integrating both sides of \eqref{eq:24} over \([0, t]\), leads to
\begin{equation} \label{eq:11}
\begin{aligned}
V(t) \le& e^{-c_1 t} \int_0^t (\kappa N(\zeta (t)) + 1) \dot{\zeta}(t)e^{c_1 \tau} d\tau\\
&+ \left( V(0) - \frac{c_2}{c_1} \right) e^{-c_1 t} + \frac{c_2}{c_1}. 
\end{aligned}
\end{equation}
Since \(0 < e^{-c_1 t} \leq 1\), the inequality \eqref{eq:11} can be rewritten as,
\begin{equation} \label{eq:25}
V(t) \le e^{-c_1 t} \int_0^t (\kappa N(\zeta (t)) + 1) \dot{\zeta}(t)e^{c_1 \tau} d\tau + c_0, 
\end{equation}
where, \(c_0 = \frac{c_2}{c_1} + V(0)\) is a positive constant. Utilizing Lemma 2 we can conclude from \eqref{eq:25} that $V(t)$, $\zeta(t)$, and $\int_{0}^{t} \left( \kappa N(\zeta (t)) + 1 \right) \dot{\zeta}(t) \tau d\tau$ are bounded on $[0, t_f)$. Considering the Lyapunov function \eqref{Lyap}, it holds that $\Psi(t)$ and $\Tilde{\psi}(t)$ 
are bounded. Thus, since ${\psi}^*(t)$ is bounded, then, $\hat{\psi}(t)$ is bounded. In addition, using Lemma 1, the boundedness of $\Psi(t)$ ensures that $e(t)$, $\dot{e}(t)$, and $\int_{0}^{t} e(\cdot)d\tau$ are bounded. Then, considering Assumption 2,
the boundedness of $e(t)$, and $\dot{e}(t)$, ensures $q(t)$, and $\dot{q}(t)$  are bounded. Furthermore, considering \eqref{kappa}, \eqref{eq:hat{psi}}, \eqref{eq:dot{zeta}}, and \eqref{eq:simplified_control},  and considering the boundedness of the basis function vector $\phi(x)$  (see \cite{nohooji2024actor} for the reference), and the boundedness of $\Psi(t)$, $\hat{\psi}(t)$,  then  $\kappa_{\Delta}(t)$, ${\psi}(t)$, $\zeta(t)$, and control $u(t)$ are bounded. Finally, as $u(t)$ is bounded, the boundedness of $\tau(t)$ is ensured, and accordingly, all closed-loop signals are bounded.

\noindent
To prove that the error $e(t)$ converges to a small neighborhood of zero, considering the boundedness of $V(0)$ and $\int_{0}^{t} \left( \kappa N(\zeta(t)) + 1 \right) \dot{\zeta}(t) \tau d\tau$, along with Properties 1 and 3, it is concluded from \eqref{Lyap} and \eqref{eq:25} that 
\[
\lim_{t \to \infty} \frac{1}{2} \|\Psi(t)\|^2 \leq \frac{c_2}{c_1},
\]
follows by $\|\Psi(t)\| \le \sqrt{\frac{2 c_2}{c_1}} := \Psi_m$. Then, considering \eqref{eq:E}, as detailed in \cite{nohooji2020constrained}, it is concluded that the error $e(t)$ converges to close to zero.
\end{proof}

%% file: Robot/05-Simulation.tex
\section{Numerical Simulation}
\label{sec_simulation}

\begin{figure}[htbp]
\centerline{\includegraphics[width=\linewidth]{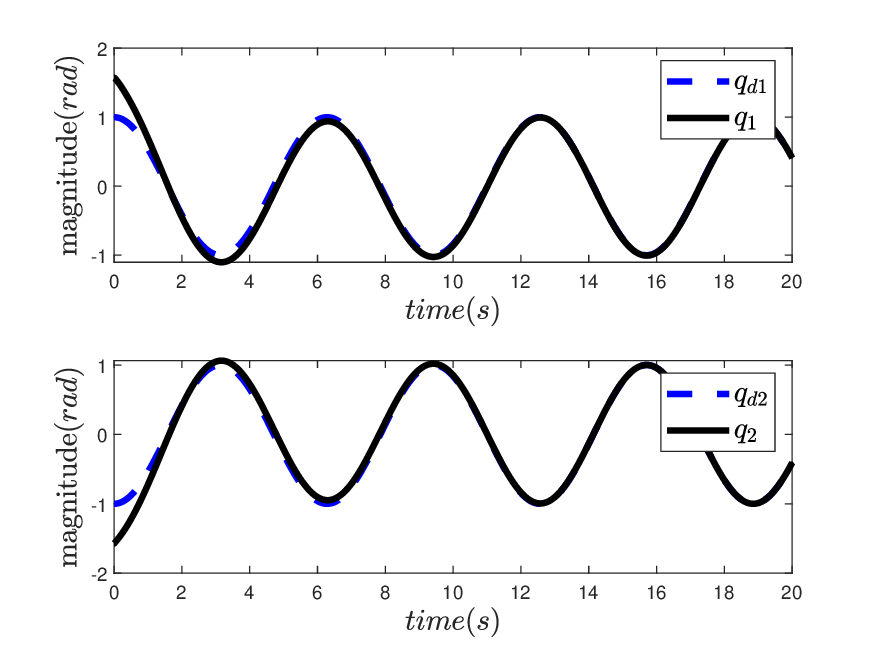}}
\caption{Desired and actual trajectories of joint positions}
\label{fig:figure1}
\end{figure}

\begin{figure}[htbp]
\centerline{\includegraphics[width=\linewidth]{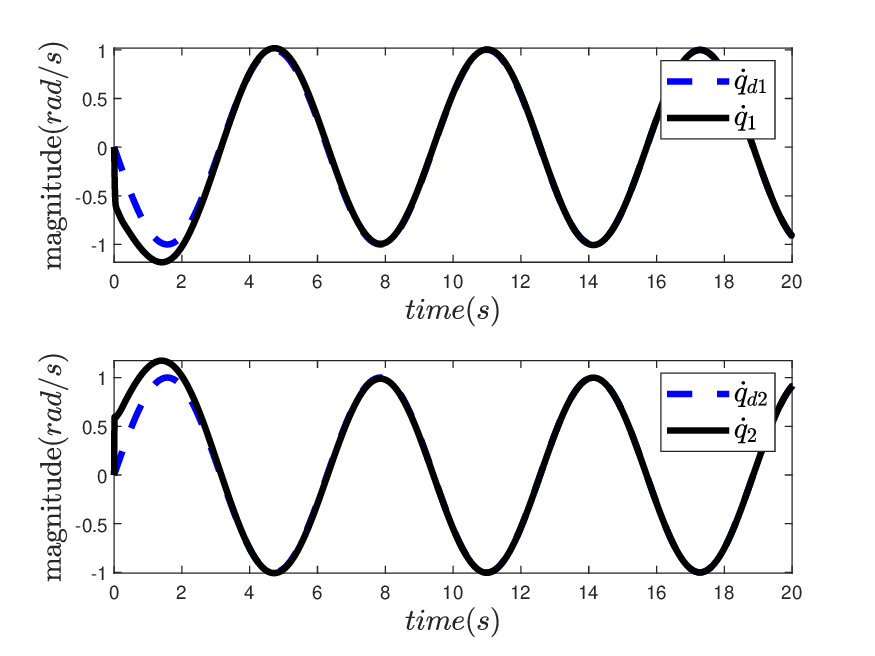}}
\caption{Desired and actual trajectories of joint velocities}
\label{fig:figure2}
\end{figure}

\begin{figure}[htbp]
\centerline{\includegraphics[width=\linewidth]{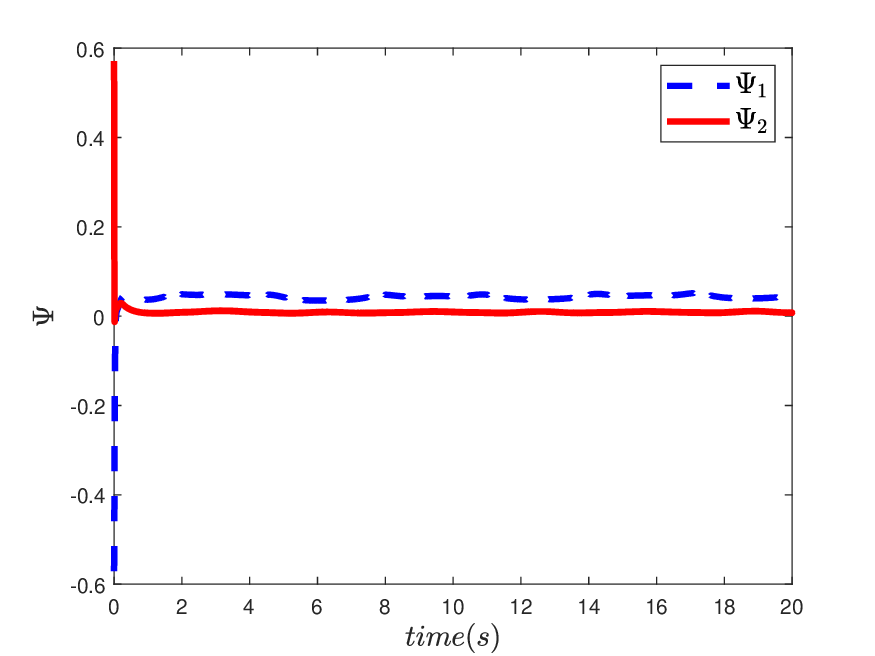}}
\caption{Trajectories of the filtered error $\Psi(t)$}
\label{fig:Psi}
\end{figure}

\begin{figure}[htbp]
\centerline{\includegraphics[width=\linewidth]{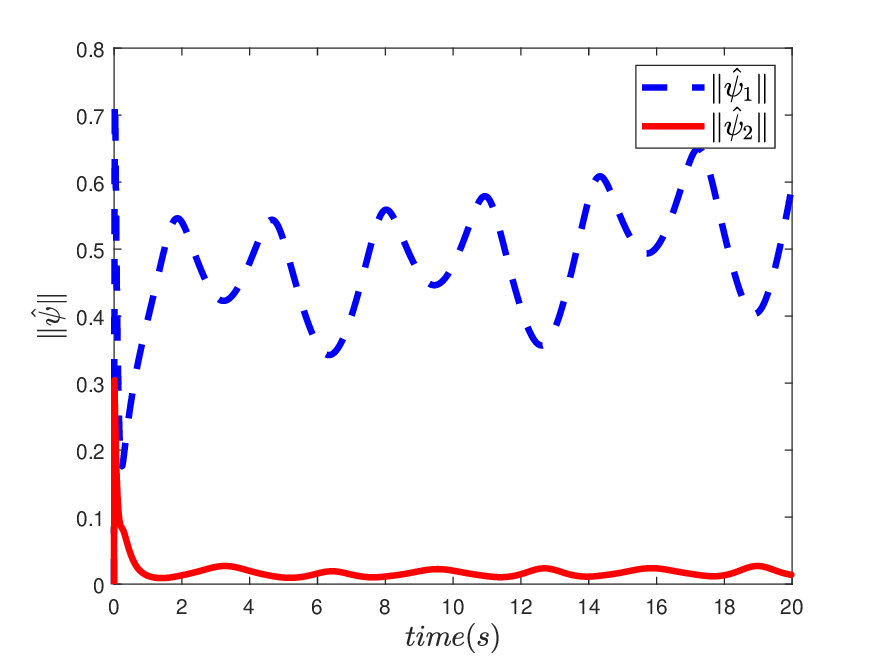}}
\caption{Norms of neural networks weignts}
\label{fig:rbf}
\end{figure}

\begin{figure}[htbp]
\centerline{\includegraphics[width=\linewidth]{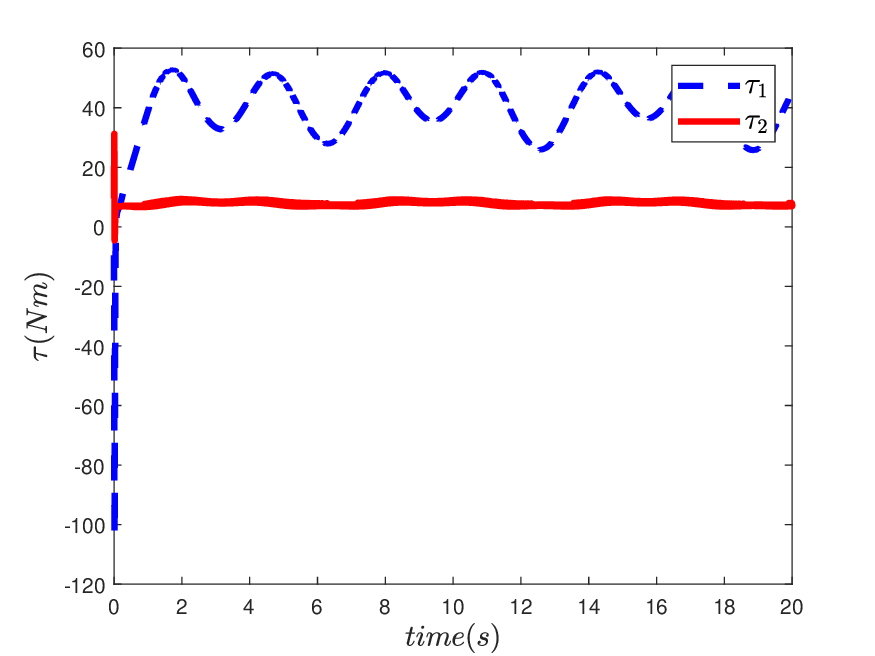}}
\caption{Trajectories of control input}
\label{fig:control}
\end{figure}

In this section, numerical simulations are performed to verify the effectiveness of the proposed Nussbaum-based PID control, as established in Theorem 1. We use a two-link robot manipulator situated in the vertical plane for our simulation study. 
Physical parameters of the robot were selected as follows: the masses of the links \(m_1 = 5\  \text{kg},\  m_2 = 2\ \text{kg}\), the lengths of the links \(l_1 = 1\  \text{m},\ l_2 = 0.75\ \text{m}\), and the inertias of the links \(I_1 = 1.66\ \text{m}^2,\ I_2 = 0.37\ \text{m}^2\). The reference trajectories are chosen as \(q_d = [\cos(t);\ -\cos(t)]\), with the initial conditions for each joint given by \(q(0) = [\pi/2; -\pi/2]\), \ and \(\dot{q}(0) = [0; 0]\).  Control parameters are chosen to be \(\alpha = 100\), \(\Gamma = 100\), \(k_\Delta = 0.1\), \(\gamma = 0.5\), and \(\sigma = 0.1\). Also, a radial basis function neural network with twenty nodes in each hidden layer is selected so that centers are evenly distributed in the span of the input space \([-12.5, 12.5]\), and widths 1.
The input vector of the neural network is chosen as
$x=\left[e^T,\dot{e}^T,q^T,\Psi^T\right]$.
The initial points of neural network weights were chosen as $\hat{\psi } \left(0\right)=0$.  Simulation results are shown in Figures \ref{fig:figure1}-\ref{fig:control}. In these figures, indices 1 and 2 denote the first and second links of the robot manipulator, respectively.

The tracking performances of the links are depicted in Figures \ref{fig:figure1}- \ref{fig:Psi}. Figures \ref{fig:figure1}, and \ref{fig:figure2} illustrate that the actual position and velocity signals closely follow their desired trajectories. 
Figure \ref{fig:Psi} illustrates the boundedness of the PID
generalized error. 
Figure \ref{fig:rbf} shows the boundedness of neural network weights.
Finally, the input control is depicted in Figure \ref{fig:control}. The above figures show the tracking performance using our developed control. It also shows that the input control vectors are bounded, demonstrating the proposed method's capability to accomplish the control tasks effectively.

% \begin{figure}[h]%\centering

% \centering
%     \includegraphics[width=0.85\textwidth]{Sections/Figs/Tau.eps}
% 	\caption{Trajectories of control input }\label{FIG_Tau}

% \end{figure}

%% file: Robot/07-Conclusion.tex
\section{Conclusion}
\label{sec_conclusion}

We presented a novel control strategy for robotic manipulators, leveraging a Nussbaum function-based PID approach to address unknown control directions and dynamics. This strategy simplifies the control design by minimizing the number of tuning parameters and utilizing direct Lyapunov analysis for stability assurance. Our contributions include a simple yet effective PID control framework, automatic gain adjustment through neural network-based estimation, and a reduction in parameter tuning complexity, enhancing adaptability and robustness in uncertain environments. 
The analytical and numerical validations highlight the effectiveness of our approach in enhancing robotic manipulator control.
These findings lay a robust groundwork for future investigations in this field, setting the stage for experimental verification and the exploration of further control innovations.

%% file: Main.bbl
\begin{thebibliography}{10}
\providecommand{\url}[1]{#1}
\csname url@samestyle\endcsname
\providecommand{\newblock}{\relax}
\providecommand{\bibinfo}[2]{#2}
\providecommand{\BIBentrySTDinterwordspacing}{\spaceskip=0pt\relax}
\providecommand{\BIBentryALTinterwordstretchfactor}{4}
\providecommand{\BIBentryALTinterwordspacing}{\spaceskip=\fontdimen2\font plus
\BIBentryALTinterwordstretchfactor\fontdimen3\font minus \fontdimen4\font\relax}
\providecommand{\BIBforeignlanguage}[2]{{%
\expandafter\ifx\csname l@#1\endcsname\relax
\typeout{** WARNING: IEEEtran.bst: No hyphenation pattern has been}%
\typeout{** loaded for the language `#1'. Using the pattern for}%
\typeout{** the default language instead.}%
\else
\language=\csname l@#1\endcsname
\fi
#2}}
\providecommand{\BIBdecl}{\relax}
\BIBdecl

\bibitem{yilmaz2021self}
B.~M. Yilmaz, E.~Tatlicioglu, A.~Savran, and M.~Alci, ``Self-adjusting fuzzy logic based control of robot manipulators in task space,'' \emph{IEEE Transactions on Industrial Electronics}, vol.~69, no.~2, pp. 1620--1629, 2021.

\bibitem{lightcap2009extended}
C.~A. Lightcap and S.~A. Banks, ``An extended kalman filter for real-time estimation and control of a rigid-link flexible-joint manipulator,'' \emph{IEEE Transactions on Control Systems Technology}, vol.~18, no.~1, pp. 91--103, 2009.

\bibitem{zhu2020estimation}
M.~Zhu, L.~Ye, and X.~Ma, ``Estimation-based quadratic iterative learning control for trajectory tracking of robotic manipulator with uncertain parameters,'' \emph{IEEE Access}, vol.~8, pp. 43\,122--43\,133, 2020.

\bibitem{nohooji2024actor}
H.~Rahimi~Nohooji, A.~Zaraki, and H.~Voos, ``Actor--critic learning based pid control for robotic manipulators,'' \emph{Applied Soft Computing}, vol. 151, p. 111153, 2024.

\bibitem{cervantes2001pid}
I.~Cervantes and J.~Alvarez-Ramirez, ``On the pid tracking control of robot manipulators,'' \emph{Systems \& control letters}, vol.~42, no.~1, pp. 37--46, 2001.

\bibitem{borase2021review}
R.~P. Borase, D.~Maghade, S.~Sondkar, and S.~Pawar, ``A review of pid control, tuning methods and applications,'' \emph{International Journal of Dynamics and Control}, vol.~9, pp. 818--827, 2021.

\bibitem{ajwad2015systematic}
S.~A. Ajwad, J.~Iqbal, M.~I. Ullah, and A.~Mehmood, ``A systematic review of current and emergent manipulator control approaches,'' \emph{Frontiers of mechanical engineering}, vol.~10, pp. 198--210, 2015.

\bibitem{armendariz2014neuro}
J.~Armendariz, V.~Parra-Vega, R.~Garc{\'\i}a-Rodr{\'\i}guez, and S.~Rosales, ``Neuro-fuzzy self-tuning of pid control for semiglobal exponential tracking of robot arms,'' \emph{Applied Soft Computing}, vol.~25, pp. 139--148, 2014.

\bibitem{belkadi2017robust}
A.~Belkadi, H.~Oulhadj, Y.~Touati, S.~A. Khan, and B.~Daachi, ``On the robust pid adaptive controller for exoskeletons: A particle swarm optimization based approach,'' \emph{Applied Soft Computing}, vol.~60, pp. 87--100, 2017.

\bibitem{azeez2023optimization}
M.~I. Azeez, A.~Abdelhaleem, S.~Elnaggar, K.~A. Moustafa, and K.~R. Atia, ``Optimization of pid trajectory tracking controller for a 3-dof robotic manipulator using enhanced artificial bee colony algorithm,'' \emph{Scientific reports}, vol.~13, no.~1, p. 11164, 2023.

\bibitem{huang2018tuning}
C.~Huang and C.~B. Yu, ``Tuning function design for nonlinear adaptive control systems with multiple unknown control directions,'' \emph{Automatica}, vol.~89, pp. 259--265, 2018.

\bibitem{scheinker2012minimum}
A.~Scheinker and M.~Krsti{\'c}, ``Minimum-seeking for clfs: Universal semiglobally stabilizing feedback under unknown control directions,'' \emph{IEEE Transactions on Automatic Control}, vol.~58, no.~5, pp. 1107--1122, 2012.

\bibitem{psillakis2016consensus}
H.~E. Psillakis, ``Consensus in networks of agents with unknown high-frequency gain signs and switching topology,'' \emph{IEEE Transactions on Automatic Control}, vol.~62, no.~8, pp. 3993--3998, 2016.

\bibitem{nussbaum1983some}
R.~D. Nussbaum, ``Some remarks on a conjecture in parameter adaptive control,'' \emph{Systems \& control letters}, vol.~3, no.~5, pp. 243--246, 1983.

\bibitem{zhao2022adaptive}
K.~Zhao, C.~Wen, Y.~Song, and F.~L. Lewis, ``Adaptive uniform performance control of strict-feedback nonlinear systems with time-varying control gain,'' \emph{IEEE/CAA Journal of Automatica Sinica}, vol.~10, no.~2, pp. 451--461, 2022.

\bibitem{song2017robust}
Y.~Song, X.~Huang, and C.~Wen, ``Robust adaptive fault-tolerant pid control of mimo nonlinear systems with unknown control direction,'' \emph{IEEE Transactions on Industrial Electronics}, vol.~64, no.~6, pp. 4876--4884, 2017.

\bibitem{habibi2018adaptive}
H.~Habibi, H.~Rahimi~Nohooji, and I.~Howard, ``Adaptive pid control of wind turbines for power regulation with unknown control direction and actuator faults,'' \emph{IEEE Access}, vol.~6, pp. 37\,464--37\,479, 2018.

\bibitem{habibi2019backstepping}
{Habibi, Hamed and Rahimi Nohooji, Hamed and Howard, Ian}, ``Backstepping nussbaum gain dynamic surface control for a class of input and state constrained systems with actuator faults,'' \emph{Information Sciences}, vol. 482, pp. 27--46, 2019.

\bibitem{nohooji2018neural}
H.~Rahimi~Nohooji, I.~Howard, and L.~Cui, ``Neural network adaptive control design for robot manipulators under velocity constraints,'' \emph{Journal of the Franklin Institute}, vol. 355, no.~2, pp. 693--713, 2018.

\bibitem{chen2016adaptive}
C.~Chen, Z.~Liu, Y.~Zhang, C.~P. Chen, and S.~Xie, ``Adaptive control of mimo mechanical systems with unknown actuator nonlinearities based on the nussbaum gain approach,'' \emph{IEEE/CAA Journal of Automatica Sinica}, vol.~3, no.~1, pp. 26--34, 2016.

\bibitem{chen2015saturated}
{Chen, Ci}, {Liu, Zhi}, {Zhang, Yun}, {Chen, CL Philip}, and {Xie, Shengli}, ``Saturated nussbaum function based approach for robotic systems with unknown actuator dynamics,'' \emph{IEEE transactions on cybernetics}, vol.~46, no.~10, pp. 2311--2322, 2015.

\bibitem{lewis2003robot}
F.~L. Lewis, D.~M. Dawson, and C.~T. Abdallah, \emph{Robot manipulator control: theory and practice}.\hskip 1em plus 0.5em minus 0.4em\relax CRC Press, 2003.

\bibitem{pane2019reinforcement}
Y.~P. Pane, S.~P. Nageshrao, J.~Kober, and R.~Babu{\v{s}}ka, ``Reinforcement learning based compensation methods for robot manipulators,'' \emph{Engineering Applications of Artificial Intelligence}, vol.~78, pp. 236--247, 2019.

\bibitem{chen2021tracking}
Q.~Chen, Y.~Wang, and Y.~Song, ``Tracking control of self-restructuring systems: a low-complexity neuroadaptive pid approach with guaranteed performance,'' \emph{IEEE Transactions on Cybernetics}, 2021.

\bibitem{nohooji2020constrained}
H.~Rahimi~Nohooji, ``Constrained neural adaptive pid control for robot manipulators,'' \emph{Journal of the Franklin Institute}, vol. 357, no.~7, pp. 3907--3923, 2020.

\end{thebibliography}
